\documentclass{article}

\usepackage[eandd, final]{neurips_2026}

\usepackage[utf8]{inputenc} 
\usepackage[T1]{fontenc}    
\usepackage{hyperref}       
\usepackage{url}            
\usepackage{booktabs}       
\usepackage{amsfonts}       
\usepackage{nicefrac}       
\usepackage{microtype}      
\usepackage{xcolor}         
\usepackage{graphicx}       
\usepackage{tabularx}       
\usepackage{multirow}       
\usepackage{amsmath}
\usepackage{tikz}
\usetikzlibrary{arrows.meta, positioning}
\usepackage{fvextra}
\title{Mage: Multi-Axis Evaluation of LLM-Generated Executable Game Scenes Beyond Compile-Pass Rate}

\author{%
  Hugh Xuechen Liu\\
  Chalmers University of Technology and University of Gothenburg\\
  \texttt{xuechen@chalmers.se}
  \And
  Kıvanç Tatar\\
  Chalmers University of Technology and University of Gothenburg\\
  \texttt{tatar@chalmers.se}\\
}

\begin{document}

\maketitle

\begin{abstract}
Compile-pass rate is the dominant evaluation signal for LLM code generation, yet for multi-component domain-specific artifacts it can be actively misleading. We demonstrate this on executable game scene synthesis with a four-axis evaluation protocol (named `Mage')---compile success, runtime success, structural fidelity, and mechanism adherence---applied to 858 generation attempts across four open-weight LLMs (7B--30B), 26~hand-crafted Unity goal pattern playable concepts, and two automatically extracted IR granularity levels. Direct NL-to-C\# generation achieves the highest runtime-pass rate (43\% mean) yet produces structurally vacuous scenes (mechanism $F_1 \approx 0.12$). Structural IR conditioning halves the runtime rate but recovers domain-faithful structure ($F_1$ up to 1.00). Within IR conditioning, behavior-only and full-scene granularity are statistically indistinguishable (McNemar $p = 1.0$), indicating input-level granularity saturation. These results show that compile rate is anti-correlated with functional correctness in this domain and that multi-axis evaluation is necessary to detect the divergence. We release the benchmark, replay logs, and per-record metrics for independent verification.\end{abstract}

\section{Introduction}
\label{sec:introduction}

Generating executable interactive game scenes from natural language is a frontier challenge for LLM code generation. A generated artifact must simultaneously satisfy three independent constraints: \emph{declarative structure} (which entities exist, with what components and properties), \emph{imperative logic} (how those entities respond to input and to each other), and \emph{engine-specific conventions} that govern the runtime API of a domain-specific framework such as Unity 2022 ~\citep{nystrom2014gameprogramming,gregory2018game}. Each constraint can fail independently: Code can compile while displaying nothing, render visible content while never triggering its win condition, or implement correct gameplay logic while failing at runtime instantiation. Standard code generation benchmarks evaluate pass@k against unit tests on standalone functions and offer an inadequate signal for any of these failure modes. Existing scene-generation systems either target static spatial layout without executable code~\citep{feng2024layoutgpt,hu2024scenecraft} or produce level geometry without gameplay logic~\citep{sudhakaran2023mariogpt,todd2023level}. One of the prior attempts at full executable scene synthesis conditioned LLMs on a semantic-level IR and achieved 0\% compilation across 4{,}160 attempts~\citep{liu2026grounding}, indicating that semantic-level structural grounding may be insufficient but leaving open what level \emph{is} sufficient.

This motivates two questions. First, when does compile-pass rate diverge from functional correctness, and what evaluation axes are needed to detect the divergence? Second, what factors determine whether structural grounding translates into executable, domain-faithful code? We address both by building a benchmark and running a controlled empirical study. From 26 hand-crafted Unity mini-games spanning distinct goal patterns, we automatically extract IRs at two granularity levels - behavior-only (script logic and inspector values) and full-scene (adding passive physics objects and camera) - and evaluate four open-weight code-generation models (7B--30B) across three seeds, yielding 858 generation attempts. Prompt template, API mapping rules, and the ground-truth corpus are held constant across all conditions. Only the IR level, model identity, and random seed vary. Generation quality is assessed along four axes: compile success, runtime success, structural fidelity (4 sub-dimensions), and mechanism adherence (6 sub-dimensions) - a multi-axis evaluation protocol designed to distinguish failures invisible to compile-rate-only measurement.

Our results reveal a systematic compile-correctness divergence. Direct NL-to-C\# generation achieves the highest runtime-pass rate (43\% mean across models, up to 80\% for the strongest model), yet produces structurally and behaviorally vacuous scenes (mechanism F1 of 0.12), indistinguishable from a no-trace baseline. Conditioning on a behavior-level IR drops the runtime-pass rate to 14\% but raises mechanism F1 to 0.82--1.00. Within IR conditioning, behavior-only and full-scene IR are statistically indistinguishable on every axis (McNemar $p = 1.0$ across all four models), suggesting that hand-designed input-level granularity has saturated. Together, these findings show that the compile rate is not aligned with (and in this setting, is anti-correlated with) domain-specific functional correctness.

\paragraph{Contributions.}
\begin{enumerate}
    \item \textbf{Evaluation methodology.} A four-axis protocol (compile  success, runtime success, structural fidelity, mechanism adherence) that exposes a compile-correctness divergence invisible to compile-rate-only evaluation. We show this divergence is systematic across four open-weight LLMs and three conditioning levels.
    \item \textbf{Benchmark and dataset.} 26 Unity goal patterns with two automatically extracted IR granularity levels, 858 generation outputs, and full replay logs released for independent verification.
    \item \textbf{Empirical characterization.} An initial controlled study on one engine and four open-weight LLMs, revealing (a) compile-correctness divergence between no-IR and IR conditioning, and (b)  within-IR granularity saturation (McNemar $p = 1.0$ across all four models).
    \item \textbf{Three-factor decomposition.} A framing lens (domain completeness, API-mapping adequacy, LLM execution fidelity) for interpreting when and why structural grounding fails to produce correct code.
\end{enumerate}

\section{Related work}
\label{sec:related}

Standard LLM code generation benchmarks evaluate pass@k (especially regarding compile-pass) against unit tests on standalone functions. HumanEval~\citep{chen2021evaluating} and APPS~\citep{hendrycks2021measuring} assess function-level synthesis against hidden test cases. SWE-bench~\citep{jimenez2023swe} extends this to repository-scale issue resolution but quality is still judged by whether a fixed test suite passes. These benchmarks provide no signal for multi-component artifacts where compilability and functional correctness can diverge.

LLMs have been applied to NL-to-scene generation tasks including 3D spatial layout~\citep{feng2024layoutgpt,sun2023threedgpt}, scripted scene construction~\citep{hu2024scenecraft}, and structured 3D scene description~\citep{avetisyan2024scenescript}. These works produce visual arrangements or spatial descriptions but do not generate runtime-executable code targeting a specific engine API. LLM-based generation has also been applied directly to game content, including 2D level generation~\citep{sudhakaran2023mariogpt,todd2023level} as part of a broader procedural content generation literature~\citep{shaker2016procedural}. Most relevant to our setting is~\citep{liu2026grounding} which conditioned LLMs on a domain-driven semantic-level IR for Unity scene synthesis. That study reported 0\% compilation across thousands of attempts spanning two open-source models, establishing a non-trivial floor that motivates the granularity question we study.

Evaluation of generated game content has largely relied on playability heuristics or human judgment. Level-generation work typically measures tile-level statistics (e.g., platform reachability) rather than scene-level structural or behavioral fidelity~\citep{sudhakaran2023mariogpt,todd2023level}. No existing benchmark provides multi-axis automated evaluation for LLM-generated executable game scenes. We propose such a benchmark and evaluation protocol.

Conditioning LLMs with structured knowledge has been explored at multiple levels. At the input level, in-context schema injection has been shown to improve structured output~\citep{mialon2023augmented}. Soft prompts and prefix tuning~\citep{lester2021power,li2021prefix} offer weight-level alternatives; broader PEFT methods are surveyed in~\citep{han2024parameter}. At the decoding level, grammar-constrained generation enforces structural validity~\citep{geng2023grammar,scholak2021picard,willard2023efficient}.

\section{Benchmark Construction}
\label{sec:bench}

\subsection{Game Design Patterns and Goal Playable Concept}

The benchmark comprises 26 Unity 2D mini-games covering distinct goal patterns, sourced from the publicly available Goal Playable Concept (GPC) corpus~\citep{lyu2023goal}. In game design domains, game design patterns structure players' interaction with the game system and with other players in the game. Among them, goal patterns~\citep{bjork2005patterns} formalize player-objective relationships with explicit win/lose conditions, enabling automated structural and behavioral evaluation without requiring expert judgment on every generated output. Patterns span diverse gameplay mechanics including resource collection (Ownership, Collection), competitive play (Race, Elimination), spatial reasoning (Alignment, Connection), and stealth/evasion (Stealth, Conceal). Each of these 26 goal patterns is implementable as a single-screen 2D mini-game, making it tractable for automated IR extraction and amenable to LLM-scale code generation. Details can be found in Appendix~\ref{app:gpc}.

\subsection{Ground-truth scenes}

Each goal pattern is realized as a complete Unity 2022.2 scene including prefab assets, MonoBehaviour scripts, and scene-level configuration (camera, boundaries, win/lose triggers). Scenes are sourced from the GPC corpus~\citep{lyu2023goal}. Prior to inclusion in the benchmark, we apply a one-time prefab-flattening step in the Unity Editor that expands every \texttt{PrefabInstance} into its constituent \texttt{GameObject}, \texttt{Transform}, and \texttt{MonoBehaviour} blocks, eliminating cross-file fileID resolution and simplifying downstream IR extraction (details in Appendix~\ref{app:flattening}). Each flattened scene was manually verified by playthrough to remain functionally equivalent to its original.

Ground-truth scenes serve a dual role. Their structure is the source from which IRs are automatically extracted (Section~\ref{sec:ir_pipeline}), and the parsed scene serves as the comparison target for structural fidelity and mechanism adherence metrics (Section~\ref{sec:eval}). Scene structure is exported in two forms: the flattened Unity \texttt{.unity} scene file and a lossless parsed JSON representation (\texttt{\_parsed.json}) encoding the full scene graph (component types, inspector field values, script bindings, and condition chains). Released ground-truth scenes (flattened Unity \texttt{.unity} format and parsed JSON) accompany the benchmark.

\subsection{IR extraction pipeline}
\label{sec:ir_pipeline}

IRs are extracted from each ground-truth scene by a five-step automated pipeline shared across both granularity levels.
\begin{enumerate}
    \item \textbf{Step 1 (Parse).} Parse the flattened Unity scene YAML into a lossless JSON representation (\texttt{\_parsed.json}).
    \item \textbf{Step 2 (Demand manifest).} Static analysis of the \texttt{.cs} scripts identifies which scene data each script references (inspector fields, component queries, tag lookups). For our specific datasets, all gameplay behavioral components are \texttt{.cs} scripts and vice versa.
    \item \textbf{Step 3 (Targeted extraction).} Query \texttt{\_parsed.json} for the demanded data - inspector field values, physics component configuration, prefab references.
    \item \textbf{Step 4 (Behavior extraction).} Extract method bodies from gameplay scripts and trace win/lose condition paths through inter-script call graphs.
    \item \textbf{Step 5 (Scene extraction).} Extract component data for GameObjects in the scene - scope determined by the IR granularity level.
\end{enumerate}

Extracted IRs are then sanitized to remove project-specific non-portable data (UI dependencies, sorting-layer IDs, redundant Transform components). Details are in Appendix~\ref{app:sanitization}. Figure~\ref{fig:ir_pipeline} summarizes the data flow.

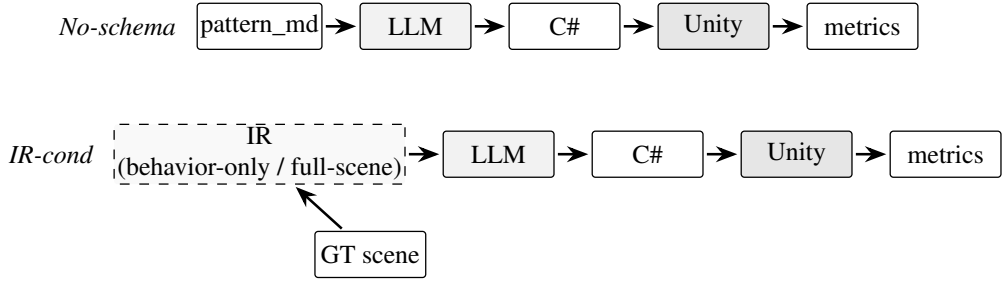
\begin{figure}[!t]
\centering
\resizebox{0.95\linewidth}{!}{%
\begin{tikzpicture}[
  node distance=2mm and 4mm,
  box/.style={draw, rounded corners=1pt, align=center, minimum height=5mm,
minimum width=12mm, inner sep=1pt, font=\footnotesize},
  data/.style={box},
  model/.style={box, fill=gray!10},
  artifact/.style={box, fill=gray!20},
  ir/.style={box, dashed, fill=gray!5},
  >={Stealth[length=1.4mm]},
  flow/.style={->, >=Stealth, shorten >=1pt, shorten <=1pt, thick},
  inject/.style={->, >=Stealth, dashed, shorten >=1pt, shorten <=1pt},
]
\node[data] (B1) {pattern\_md};
\node[model, right=of B1] (B2) {LLM};
\node[data, right=of B2] (B3) {C\#};
\node[artifact, right=of B3] (B4) {Unity};
\node[data, right=of B4] (B5) {metrics};
\node[left=1mm of B1, font=\footnotesize\itshape, anchor=east] {No-schema};
\draw[flow] (B1) -- (B2); \draw[flow] (B2) -- (B3); \draw[flow] (B3) --
(B4); \draw[flow] (B4) -- (B5);

\node[ir, below=8mm of B1] (I1) {IR\\(behavior-only / full-scene)};
\node[model, right=of I1] (I2) {LLM};
\node[data, right=of I2] (I3) {C\#};
\node[artifact, right=of I3] (I4) {Unity};
\node[data, right=of I4] (I5) {metrics};
\node[left=1mm of I1, font=\footnotesize\itshape, anchor=east] {IR-cond};
\draw[flow] (I1) -- (I2); \draw[flow] (I2) -- (I3); \draw[flow] (I3) --
(I4); \draw[flow] (I4) -- (I5);

\node[data, below=5mm of I1, xshift=12mm] (GT) {GT scene};
\draw[flow] (GT) -- (I1);
\end{tikzpicture}%
}
\caption{Two evaluation conditions. \textit{No-schema}: the LLM receives
only the natural-language pattern description. \textit{IR-cond}: the LLM
receives a ground-truth-derived IR at one of two granularity levels
(behavior-only or full-scene), extracted offline from the corresponding
Unity scene by an automated pipeline (Section~\ref{sec:ir_pipeline}); the
natural-language description is not provided. Both conditions feed into the
same Unity batch-replay harness for metric collection. Box shading: data
(white), LLM (light grey), runtime artifact (dark grey), schema (dashed).}
\label{fig:ir_pipeline}
\end{figure}

We compare two granularity levels (Table~\ref{tab:ir_levels}): \textbf{Behavior-only IR} includes script class definitions with complete method bodies, inspector field values with their types, the physics component configuration referenced by scripts, win/lose condition chains, prefab references, and gameplay-relevant tags. It omits Stage 5 entirely (e.g., no GameObject visual or scene-layout data is included). \textbf{Full-scene IR} strictly extends behavior-only by executing Stage 5 over \emph{all} GameObjects in the scene - scripted and unscripted alike - capturing camera configuration, boundary geometry, passive obstacles, and rendering components. The two levels share Step 1 and Step 2-4. They differ only in whether Step 5 is executed.

\begin{table}[t]
  \caption{IR granularity levels. Both levels share Step 1 and Step 2-4. Full-scene additionally executes Step 5 over all GameObjects in the scene.}
  \label{tab:ir_levels}
  \centering
  \small
  \begin{tabular}{lll}
    \toprule
    Level & Pipeline coverage & GameObject scope (Step 5) \\
    \midrule
    Behavior-only & Step 1 + Step 2-4 & - (Step 5 not executed) \\
    Full-scene    & Step 1 + Step 2-5 & all GameObjects \\
    \bottomrule
  \end{tabular}
\end{table}

These two levels are deliberately chosen as the structural endpoints of an enumerable design space rather than as a continuous spectrum. Behavior-only is the minimal configuration that captures gameplay logic (i.e., without it, no win/lose chain is reproducible). Full-scene is the maximal configuration short of including ground-truth runtime state, capturing all serialized scene structure but no per-frame state. Intermediate granularities (e.g., behavior plus visual components on scripted GameObjects only) can be defined within the same pipeline by parameterizing Step 5's scope but are left to follow-up work. Both IRs are released as JSON for each of the 26 patterns, accompanying the benchmark.

\subsection{Prompt templates}
\label{sec:prompts}

Three prompt templates are constructed for the three conditions (no-IR, behavior-only, full-scene). All three share an identical \emph{API mapping rule block}; they differ only in the IR content injected and the instruction governing how to use that content. Each template instructs the LLM to emit a single C\# file implementing a \emph{runtime builder pattern}: a \texttt{GameBuilder} \texttt{MonoBehaviour} whose \texttt{Awake()} method programmatically constructs the scene at runtime, with all gameplay \texttt{MonoBehaviour} classes defined inline in the same file, keeping the entire artifact within Unity's runtime assembly. When an IR is provided, it is supplied as ground-truth-derived JSON. The LLM is not asked to generate the IR itself. This isolates the code-generation step and removes IR generation as a confound.

Under all three conditions, the prompt substitutes \texttt{<PATTERN\_MD>} (no-schema) or \texttt{<IR\_JSON>} (IR conditions) with the full source verbatim. For no-schema, this is the complete pattern markdown including Overview, Examples, and Relations sections (\textasciitilde 5--10\,KB per pattern). Wikilinks (e.g., \texttt{[Capture](Capture.md)}) are preserved as-is. These rules encode common compile-time and runtime pitfalls when mapping serialized scene data to Unity's runtime API. All three templates share six API mapping rules addressing common Unity 2022 runtime pitfalls. Full prompts and rationale appear in Appendix~\ref{app:prompts}.

Crucially, the API mapping rules are held constant across all three conditions. Differences in generated code thus reflect differences in the IR content provided to the model, not differences in the prompt's engineering quality.

\section{Evaluation Protocol of Multi-Axis Game Evaluation: `Mage'}
\label{sec:eval}

Generation quality is assessed along four axes: compile success, runtime success, structural fidelity, and mechanism adherence. Compile and runtime success measure whether the generated code can be loaded and executed by Unity. Structural fidelity and mechanism adherence measure how closely the resulting scene matches the ground-truth pattern. We use these descriptive axis names throughout the paper. The released harness uses the labels \texttt{m1\_compile}, \texttt{m1\_exec}, \texttt{m2\_*}, and \texttt{m4\_*} for backward compatibility with the implementation; the full mapping appears in Appendix~\ref{app:metrics_mapping}. The four axes are defined in Sections~\ref{sec:eval_runtime}, \ref{sec:eval_struct}, and \ref{sec:eval_mech}; statistical procedures appear in Section~\ref{sec:eval_stats}.

\subsection{Compile and runtime success}
\label{sec:eval_runtime}

\textbf{Compile success.} The generated C\# file is loaded by Unity 2022.2.23f1 in batch mode. A record is counted as a compile pass if the Unity log contains no \texttt{error CS\textbackslash d+} messages, excluding \texttt{CS2001} stale-reference messages produced by the Editor's pre-write file lookup.

\textbf{Runtime success.} A record is counted as a runtime pass if it is a compile pass \emph{and} the Unity log contains the markers \texttt{GameBuilder.Awake(): OK} and a clean-exit signature. This indicates that the runtime builder executed without crashing. Runtime success is a strict subset of compile success. Both markers are emitted by the harness's batch runner script. Raw replay logs are released alongside the benchmark for independent verification.

\subsection{Structural fidelity}
\label{sec:eval_struct}

Structural fidelity measures whether the generated scene contains the GameObjects, components, scripts, and tags expected by the ground-truth pattern. We report four sub-dimensions, each computed as a multiset $F_1$ between the ground-truth scene and the parsed generated scene:
\begin{itemize}
    \item \textbf{Scripts $F_1$}: class names defined in the generated code vs.\ those declared in the ground-truth manifest.
    \item \textbf{GameObject names $F_1$}: \texttt{m\_Name} multiset over all GameObjects in the parsed scene.
    \item \textbf{Component types $F_1$}: count of each Unity component type, excluding scene-settings blocks (\texttt{OcclusionCullingSettings}, \texttt{RenderSettings}, \texttt{LightmapSettings}, \texttt{NavMeshSettings}).
    \item \textbf{Tags $F_1$}: \texttt{m\_TagString} multiset, excluding the default \texttt{Untagged} tag.
\end{itemize}
Note that structural fidelity is defined only for runtime-pass records (those that produced an executable scene). The denominator $n$ varies by condition and is reported alongside each result.

\subsection{Mechanism adherence}
\label{sec:eval_mech}

Mechanism adherence measures whether the generated code reproduces the ground-truth gameplay mechanism: the chain of events, conditions, and effects constituting the win or lose path. The metric operates by static analysis of the generated C\# source and requires no runtime execution, so it is defined for all 858 generation attempts.
\begin{itemize}
    \item \textbf{Steps $F_1$ (win/lose)}: set $F_1$ over $(actor\_class, event)$ pairs, where each step represents a Unity callback (e.g., \texttt{OnTriggerEnter2D} on \texttt{Player}) on the win or lose path.
    \item \textbf{Effects $F_1$ (win/lose)}: set $F_1$ over the effect strings produced at each step (e.g., \texttt{Destroy(gameObject)}, \texttt{GameManager.Win()}).
    \item \textbf{Conditions $F_1$ (win/lose)}: set $F_1$ over guard expressions (e.g., \texttt{other.tag == "Player"}, \texttt{score >= 10}).
\end{itemize}
Win paths are defined for all 26 patterns. Lose paths exist for 10 patterns. For the remaining 16 patterns, lose-path scores are recorded as \texttt{null} and excluded from aggregation. Path extraction parses C\# class bodies via brace-counting; method bodies that include unbalanced braces inside string literals or comments are a known blind spot.

\subsection{Statistical evaluation}
\label{sec:eval_stats}

We report pass@$k$ for $k \in \{1, 2, 3\}$ using the unbiased estimator of~\citet{chen2021evaluating}: with $n = 3$ seeds per (model, condition, pattern), per-pattern pass@$k$ is computed and macro-averaged across the 26 patterns. Pass@$k$ is defined on runtime success.

Pairwise comparisons between conditions use McNemar's exact test on per-pattern any-pass aggregation~\citep{dietterich1998approximate,mcnemar1947note}. A pattern is counted as a pass for a (model, condition) cell if any of its three seeds achieves runtime success. Each (model, pattern) thus contributes one paired observation. We report the exact two-sided $p$-value, the discordance odds ratio with a 95\% Woolf CI, and Cohen's $h$.

All raw replay logs and per-record metric values are released alongside the benchmark. The released harness reproduces every reported number from these logs without re-running generation or replay.

\section{Empirical Study}
\label{sec:study}

\subsection{Setup}
\label{sec:study_setup}

We evaluate four open-weight code-and-general LLMs spanning 7B to 30B active parameters across two IR granularity levels and a no-IR baseline. Each (model, condition, pattern) cell is generated with three random seeds. Given the 26 patterns, there are theoretically 936 attempts. However, one of the (model, condition) scenarios has been excluded, resulting in 858 total generation attempts as a result - namely (4 models x 3 conditions - 1) x 26 patterns x 3 seeds. Details are given in the following paragraph.

The four models are Qwen3-Coder-30B-A3B-Instruct~\citep{qwen3technicalreport} (30B mixture-of-experts, 3B active parameters; code-specialized, 2025), Gemma-4-26B-A4B-it~\footnote{\url{https://huggingface.co/google/gemma-4-26B-A4B-it}} (26B mixture-of-experts, 4B active; general-purpose, 2026), DeepSeek-Coder-V2-Lite-Instruct~\citep{deepseekai2026deepseekv4} (16B; code-specialized, 2024), and Qwen2.5-Coder-7B-Instruct~\citep{hui2024qwen2,qwen2} (7B; code-specialized, 2024). All four fit on a single H100 80GB in BF16 with vLLM. Note DeepSeek-Coder-V2-Lite and Qwen2.5-Coder-7B are the same models used in fore-mentioned semantic-IR work~\citep{liu2026grounding}, isolating IR effects from model scale. However, we exclude (Qwen2.5-Coder-7B, full-scene IR) due to context-window overflow on the largest patterns at 32K tokens. 

Generation uses temperature 0.7 and top-p 0.95 with a 32K-token output budget. All conditions share identical prompt templates and API mapping rules (Section~\ref{sec:prompts}). 

Code and data for this project are available at \url{https://anonymous.4open.science/r/neurips-ir-benchmark-game-scene-3F65/} and \url{https://huggingface.co/datasets/anon-neurips-2026-0502/scene-level-grounding-benchmark}

\subsection{Results}
\subsubsection{Compile-correctness divergence}
\label{sec:study_divergence}

Table~\ref{tab:m1_funnel} reports compile and runtime success across the four models and three conditions. Direct NL-to-C\# generation (no-schema) achieves the highest mean runtime-pass rate at 43\%, with Qwen3 reaching 80\% on no-schema. Behavior-only and full-scene IR conditioning reduce mean runtime success to 14\% and 21\% respectively, with the gap concentrated on stronger models: Qwen3 drops from 80\% (no-schema) to 37\% / 42\% (behavior-only / full-scene).

\begin{table}[t]
  \caption{Compile and runtime success rates per model and IR condition. C = compile success; R = runtime success. Counts in parentheses; runtime success is a strict subset of compile success. Qwen2.5 $\times$ full-scene is excluded due to 32K context overflow.}
  \label{tab:m1_funnel}
  \centering
  \small
  \begin{tabular}{l|cc|cc|cc}
    \toprule
                & \multicolumn{2}{c|}{No-schema} &
    \multicolumn{2}{c|}{Behavior-only} & \multicolumn{2}{c}{Full-scene} \\
                & C & R & C & R & C & R \\
    \midrule
    Qwen3-30B   & 83\% (65) & \textbf{80\% (63)} & 42\% (33) & 37\% (29) & 47\% (37) & 42\% (33) \\
    Qwen2.5-7B  & 74\% (58) & 39\% (31) & 3\% (3) & 3\% (3) & - & - \\
    DeepSeek-16B & 23\% (18) & 18\% (14) & 29\% (23) & 16\% (13) & 42\% (33) & 21\% (17) \\
    Gemma-4-26B & 34\% (27) & 34\% (27) & 1\% (1) & 1\% (1) & 0\% (0) & 0\% (0) \\
    \midrule
    Mean        & 54\% & 43\% & 19\% & 14\% & 29\% & 21\% \\
    \bottomrule
  \end{tabular}
\end{table}

Table~\ref{tab:beyond_compile} reports structural fidelity and mechanism adherence on the same conditions. The pattern reverses. No-schema scores collapse on every fidelity dimension: tag $F_1$ is exactly 0.000 across all four models, indicating that none of the runtime-pass scenes contain the gameplay-relevant tags expected by the ground-truth pattern. Mechanism win-step $F_1$ is approximately 0.115 across all four no-schema cells-a baseline floor reflecting condition-path tracing failure rather than gameplay correctness. With behavior-only or full-scene IR, mechanism win-step $F_1$ rises to 0.82--1.00 on the strongest model and 0.61--0.97 across all IR-conditioned cells.

\begin{table}[t]
 \caption{Structural fidelity and mechanism adherence (win-path $F_1$) for each (model, condition) cell. Structural fidelity is computed only on runtime-pass records ($n$ varies, shown in subscript); mechanism adherence is computed on all 858 attempts. Lose-path scores appear in Appendix~\ref{app:lose_path}. The constant 0.12 floor across all no-schema cells reflects three patterns (Evade, Guard, Survive) whose ground-truth win paths are empty; the $F_1(\emptyset, \emptyset) = 1.0$ convention produces $3/26 \approx 0.12$ regardless of model output. \textbf{---} indicates no data: Qwen2.5$\times$Full-scene was excluded due to context overflow; Gemma4$\times$Full-scene had zero runtime-pass records ($n{=}0$).}
  \label{tab:beyond_compile}
  \centering
  \scriptsize
  \setlength{\tabcolsep}{4pt}
  \begin{tabular}{ll|cccc|ccc}
    \toprule
    Condition & Model & Scripts & GO & Comp & Tags & Steps & Effects & Conds \\
    \midrule
    \multirow{4}{*}{Full-scene}
      & Qwen3        & 0.84$_{n=33}$ & 0.56 & 0.56 & 0.94 & 0.96 & 0.97 & 0.93 \\
      & DeepSeek     & 0.26$_{n=17}$ & 0.06 & 0.33 & 0.51 & 0.61 & 0.63 & 0.60 \\
      & Gemma4       & -           & -  & -  & -  & 0.92 & 0.98 & 0.66 \\
      & Qwen2.5      & -           & -  & -  & -  & -  & -  & -  \\
    \midrule
    \multirow{4}{*}{Behavior-only}
      & Qwen3        & 0.83$_{n=29}$ & 0.26 & 0.48 & 0.60 & 1.00 & 1.00 & 0.97 \\
      & DeepSeek     & 0.74$_{n=13}$ & 0.09 & 0.24 & 0.15 & 0.82 & 0.85 & 0.80 \\
      & Gemma4       & 0.75$_{n=1}$  & 0.20 & 0.37 & 0.80 & 0.97 & 0.98 & 0.67 \\
      & Qwen2.5      & 0.69$_{n=3}$  & 0.25 & 0.49 & 0.17 & 0.87 & 0.88 & 0.85 \\
    \midrule
    \multirow{4}{*}{No-schema}
      & Qwen3        & 0.15$_{n=63}$ & 0.07 & 0.32 & \textbf{0.00} & 0.12 & 0.12 & 0.12 \\
      & DeepSeek     & 0.03$_{n=16}$ & 0.04 & 0.19 & \textbf{0.00} & 0.12 & 0.12 & 0.12 \\
      & Gemma4       & 0.21$_{n=27}$ & 0.09 & 0.43 & \textbf{0.00} & 0.12 & 0.12 & 0.12 \\
      & Qwen2.5      & 0.02$_{n=31}$ & 0.03 & 0.19 & \textbf{0.00} & 0.12 & 0.12 & 0.12 \\
    \bottomrule
  \end{tabular}
\end{table}

\subsubsection{Within-IR granularity saturation}
\label{sec:study_saturation}

Table~\ref{tab:mcnemar} reports pairwise McNemar tests on per-pattern any-pass aggregation. Within IR conditioning, we find no statistical evidence of differences between behavior-only and full-scene at any (model, condition) pair: McNemar $p = 1.0$ across all four models. By contrast, IR-vs-no-schema comparisons are significant for three of four models on at least one IR level (Qwen3, Gemma-4, Qwen2.5; $p < 0.01$), with no-schema as the higher-runtime condition.

\begin{table}[t]
  \caption{Pairwise McNemar exact tests on per-pattern any-pass aggregation. $\text{prop}_A$ and $\text{prop}_B$ are the proportions of 26 patterns with at least one runtime-passing seed under condition $A$ and $B$ respectively. OR is the discordance odds ratio with 95\% Woolf CI; Cohen's $h$ measures effect size. \textbf{*} indicates $p < 0.05$.}
  \label{tab:mcnemar}
  \centering
  \scriptsize
  \begin{tabular}{lll|cc|c|cc|c}
    \toprule
    \multicolumn{2}{l}{Comparison} & Model & $\text{prop}_A$ & $\text{prop}_B$ & $p$
    & OR & 95\% CI & Cohen's $h$ \\
    \midrule
    Full-scene & Behavior-only & Qwen3      & 0.65 & 0.69 & 1.00 & 0.75 & [0.17, 3.35] & 0.08 \\
    Full-scene & Behavior-only & DeepSeek   & 0.39 & 0.39 & 1.00 & 1.00 & [0.32, 3.10] & 0.00 \\
    Full-scene & Behavior-only & Gemma4     & 0.00 & 0.04 & 1.00 & -  & -          & -  \\
    \midrule
    Full-scene & No-schema     & Qwen3      & 0.65 & 1.00 & \textbf{0.004*} & - & - & 1.26 \\
    Full-scene & No-schema     & DeepSeek   & 0.39 & 0.42 & 1.00 & 0.86 & [0.29, 2.55] & 0.08 \\
    Full-scene & No-schema     & Gemma4     & 0.00 & 0.54 & \textbf{0.0001*} & - & - & - \\
    \midrule
    Behavior-only & No-schema  & Qwen3      & 0.69 & 1.00 & \textbf{0.008*} & - & - & 1.18 \\
    Behavior-only & No-schema  & Qwen2.5    & 0.12 & 0.62 & \textbf{0.001*} & 0.07 & [0.01, 0.54] & 1.11 \\
    Behavior-only & No-schema  & DeepSeek   & 0.39 & 0.42 & 1.00 & 0.86 & [0.29, 2.55] & 0.08 \\
    Behavior-only & No-schema  & Gemma4     & 0.04 & 0.54 & \textbf{0.0002*} & - & - & 1.25 \\
    \bottomrule
  \end{tabular}
\end{table}

Within IR conditioning, the benchmark cannot distinguish the two granularities. We discuss implications for learned conditioning methods in Section~\ref{sec:disc}.

\subsubsection{Failure mode taxonomy}
\label{sec:study_failures}

Table~\ref{tab:errors} reports the top compile-error codes across the 858 attempts (5{,}097 total occurrences after CS2001 noise filtering). Semantic errors dominate: \texttt{CS0122} (inaccessible member, 991), \texttt{CS0029} (implicit type conversion, 730), \texttt{CS0136} (variable scope, 504), and \texttt{CS1061} (missing method, 495) account for more than half of all errors. Two condition-specific patterns are notable. \texttt{CS0246} (type/namespace not found) appears exclusively under IR conditioning (469 occurrences vs 0 in no-schema), indicating that routing IR entity names through the prompt's API-mapping rules introduces type-resolution failures that the no-schema condition entirely bypasses. CS0122 (inaccessible member) spikes under behavior-only IR (862 vs 117 full-scene), as the IR's inspector field listing surfaces access modifiers the model misapplies. We return to these patterns in Section~\ref{sec:disc}.

\begin{table}[t]
  \caption{Top 10 compile-error codes across the 858 evaluation attempts. NS = no-schema, BO = behavior-only, FS = full-scene.}
  \label{tab:errors}
  \centering
  \small
  \begin{tabular}{llrrrr}
    \toprule
    Code & Meaning & Total & NS & BO & FS \\
    \midrule
    CS0122 & Inaccessible due to protection level    & 991 & 12  & 862 & 117 \\
    CS0029 & Cannot implicitly convert type          & 730 & 219 & 291 & 220 \\
    CS0136 & Local variable name conflict (scope)    & 504 & 150 & 126 & 228 \\
    CS1061 & Type does not contain definition        & 495 & 102 & 151 & 242 \\
    CS0246 & Type/namespace name not found           & 469 &   0 & 320 & 149 \\
    CS0117 & Type does not contain definition (field)& 435 &  93 & 192 & 150 \\
    CS0103 & Name does not exist in context          & 383 &  33 & 272 &  78 \\
    CS1503 & Cannot convert argument type            & 283 &  63 & 106 & 114 \\
    CS0176 & Static member via instance reference    &  93 &   0 &  87 &   6 \\
    CS0120 & Non-static field via type reference     &  60 &  54 &   0 &   6 \\
    \bottomrule
  \end{tabular}
\end{table}

\section{Discussion}
\label{sec:disc}
\paragraph{Compile rate alone as a misleading evaluation signal.}
The compile-correctness divergence (Section~\ref{sec:study}) suggests that compile-pass rate, in this domain, is a proxy for syntactic and reference-level conformance to C\# and the Unity API surface, not for functional correctness. Without an IR, the LLM tends to emit small, locally consistent programs (empty \texttt{Awake()} bodies, single-GameObject scenes, default sprite assignments) that satisfy the parser but produce scenes containing none of the structure or behavior defining the target pattern. Conditioning on a structural IR forces a much larger code surface, each additional element a potential failure point. Compile rate measures \emph{programs the model dared to write that the parser accepted}, not \emph{programs that implement the requested artifact}. The error taxonomy in Section~\ref{sec:study_failures} corroborates this: failures concentrate at the Unity API semantics layer, not the C\# parser layer.

\paragraph{A three-factor interpretation, anchored in the error taxonomy.}
The empirical findings can be organized along three independent factors-\emph{domain completeness}, \emph{API-mapping adequacy}, and
\emph{LLM execution fidelity}—each visible in distinct corners of
Sections~\ref{sec:study} and~\ref{sec:study_failures}. \textbf{Factor 1 (domain completeness):} what information the IR carries. Direct NL$\rightarrow$C\# lacks any structural IR and produces mechanism $F_1 \approx 0.115$ across all four models as a baseline floor. Both IR levels supply Factor~1. Within-IR saturation (Section~\ref{sec:study}) shows behavior-level information already suffices. Yet richer information is not uniformly beneficial. Behavior-only IR's inspector field listings push \texttt{CS0122} (inaccessible-member errors) to 862 occurrences versus 117 under full-scene (Table~\ref{tab:errors}), as the model misapplies the access modifiers the IR makes salient. \textbf{Factor 2 (API-mapping adequacy):} whether the prompt translates IR entity names into valid Unity API calls. The rule block is held constant across conditions, but its \emph{load} differs—IR conditions require explicit IR-to-API translation that no-schema sidesteps. \texttt{CS0246} (type/namespace not found) occurs 469 times under IR conditioning and 0 times under no-schema (Table~\ref{tab:errors}). This gap is the visible cost of routing through a translation step the rules do not exhaustively cover. \textbf{Factor 3 (LLM execution fidelity):} whether the model uses the information correctly. Factor~3 dominates cross-model variance: behavior-only IR yields 37\% runtime success on Qwen3-30B but 3\% on Qwen2.5-7B and 1\% on Gemma-4-26B—a 12--37$\times$ gap on the same input. We treat this decomposition as a framing tool rather than a set of independently measured contributions. Controlled factorial interventions on each factor are needed for causal estimates.

\paragraph{Failure distribution reshapes under IR conditioning.}
Compile failures across all conditions are dominated by semantic errors-type resolution, access control, scope conflicts-not syntax errors (Section~\ref{sec:study_failures}). Models reliably produce parseable C\# but stumble on Unity API semantics. The shift from no-schema to IR conditioning reshapes the failure distribution rather than reducing it: \texttt{CS0246} and \texttt{CS0122} concentrate under IR conditions because the IR introduces type and member references the model must resolve correctly, surfacing semantic failures that no-schema generation never attempts.

\paragraph{Implications for benchmark design.}
Any LLM code-generation task targeting a heavy domain-specific framework (e.g., game engines, simulation environments, build systems) faces a similar three-factor structure. Single-axis benchmarks (compile-pass, unit-test-pass) cannot distinguish a structurally vacuous program from a complete one if both pass the parser. Such benchmarks should report at least three axes: parser-level pass, runtime-level pass, and domain-fidelity pass.

We must evaluate LLM-generated interactive code beyond compile-pass rate. Structural fidelity and mechanism adherence must be first-class metrics. We release this benchmark and harness as a step toward that evaluation regime.

\section{Limitations}
\label{sec:limitations}

\begin{itemize}
    \item \textbf{Runtime playability not measured.} Runtime success indicates only that \texttt{Awake()} executed and Unity exited cleanly, not that a human player can complete the win condition. The released harness supports playability evaluation.

    \item \textbf{Single engine.} The protocol is validated on Unity 2022.2 with 26 goal patterns from~\citep{bjork2005patterns}. Cross-engine replication (e.g., Godot, three.js) is the natural next step that would test the generality of the findings.

    \item \textbf{Sample size suitable for evaluation, not training.} 26 patterns provide statistical power for the reported comparisons but are insufficient for de-novo training. Follow-up work should plan for frozen pretrained encoders.

    \item \textbf{Limited model coverage.} Four open-weight models (7B--30B active parameters) on one hyperparameter configuration suffice to demonstrate the protocol and the compile-correctness divergence, but do not constitute an exhaustive survey. Extending to closed-source models, larger open-weight models, and specialized code agents is future work.

    \item \textbf{Three-factor decomposition is interpretive.} As stated in  Section~\ref{sec:disc}, the three factors organize observations but are not independently measured. Controlled factorial interventions are needed for causal estimates.
\end{itemize}

\section{Conclusion}
\label{sec:concl}

We present a benchmark and systematic study of structural IR conditioning for LLM-generated Unity scene code: 26 hand-crafted patterns, two granularity levels of automatically extracted IR, four open-weight LLMs, and a four-axis evaluation harness covering compile success, runtime success, structural fidelity, and mechanism adherence. Across 858 generation attempts, direct natural-language-to-C\# generation achieves the highest runtime-pass rate yet produces scenes that are structurally and behaviorally vacuous (mechanism $F_1 \approx 0.12$). Structural IR conditioning sacrifices runtime-pass rate but recovers domain-faithful structure and mechanism (mechanism $F_1$ up to $1.00$). Within IR conditioning, behavior-only and full-scene granularity are statistically indistinguishable, suggesting that hand-designed input-level conditioning has saturated for this benchmark. We release the patterns, IR extraction pipeline, prompt templates, evaluation harness, and 858 generation outputs with full replay logs and per-record metrics for independent verification and extension of these findings. The four-axis protocol is engine- and model-agnostic by design. Applying it to additional engines, model families, and generation strategies - and testing whether the compile-correctness divergence persists across these axes - is the immediate next step as future work.

\begin{ack}
This work was partially supported by the Wallenberg AI, Autonomous Systems and Software Program – Humanities and Society (WASP-HS) funded by the Marianne and Marcus Wallenberg Foundation and the Marcus and Amalia Wallenberg Foundation. 

The computations and data handling were enabled by resources provided by the National Academic Infrastructure for Supercomputing in Sweden (NAISS), partially funded by the Swedish Research Council through grant agreement no. 2022-06725.
\end{ack}

\newpage
\bibliographystyle{plainnat}
\bibliography{references}


\appendix

\section{Appendix}
\label{sec:appendix}

\subsection{Goal Playable Concepts}
\label{app:gpc}

\begin{figure}
\centering
\includegraphics[width=\linewidth]{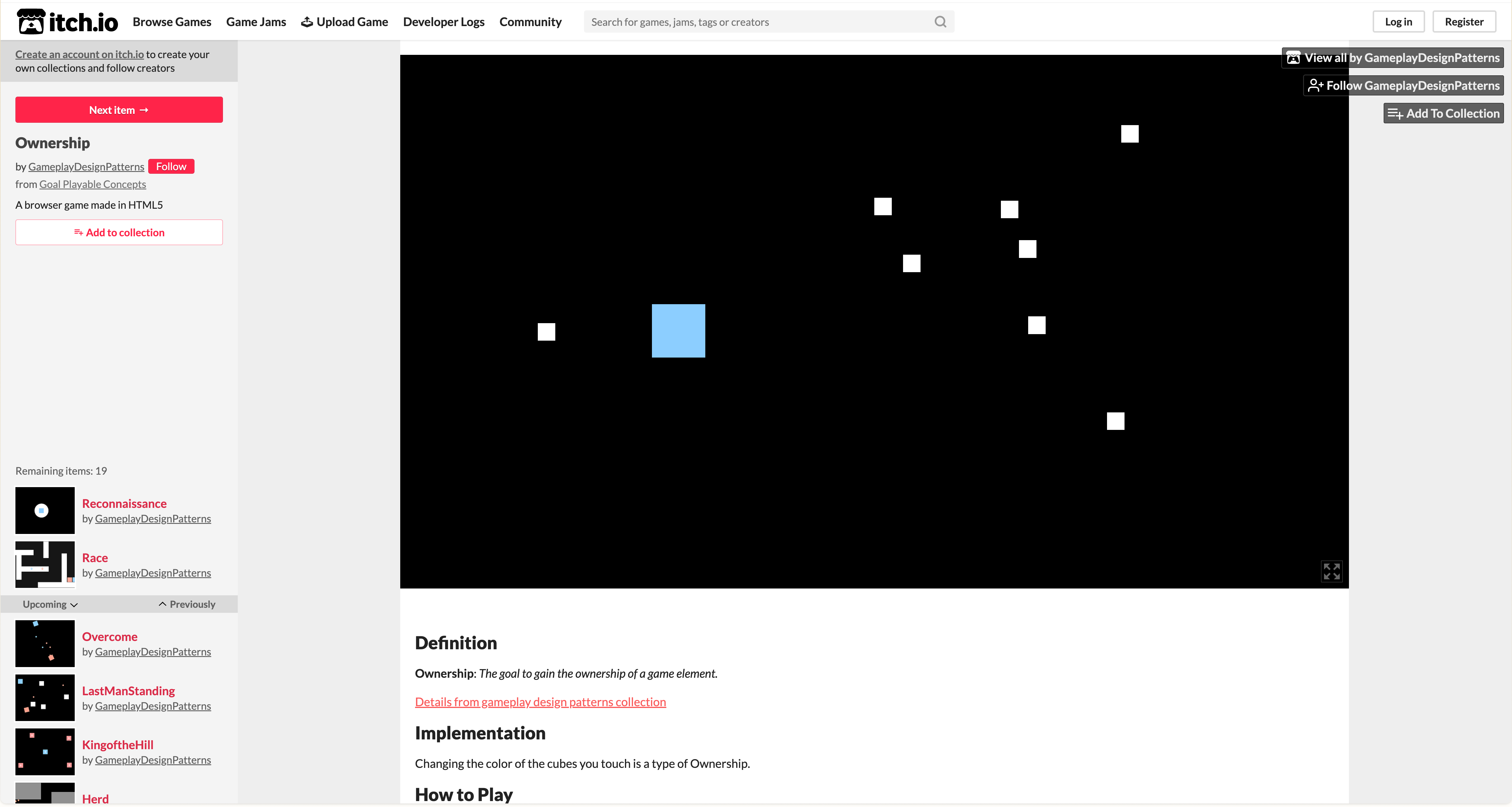}
\caption{Screenshot of the ``Ownership'' goal playable concept running in the browser. The player controls the large blue box and must collide with all small white boxes to claim ownership.}
\label{fig:ownership-gpc}
\end{figure}

The 26 goal playable concept Unity mini-games can be found at \url{https://itch.io/queue/c/3720943/goal-playable-concepts?game_id=1775436&password=}. Figure~\ref{fig:ownership-gpc} shows a screenshot of the mini-game titled ``Ownership'' running in the browser.

In this game, the initial scene contains a large blue box and multiple small white boxes. The player controls the blue box using the WASD keys. When the player collides with a small white box, it turns blue, indicating that it has been ``owned.'' The win condition is to own all eight white boxes in the scene. There is no lose condition for this particular goal pattern mini-game.

\subsection{Prompt templates}
\label{app:prompts}

This subsection reproduces the three prompt templates verbatim. All three share the same API mapping rule block (Section~\ref{sec:prompts}); they differ only in the IR content section and the instruction governing how to use it.

Notice there are six shared API mapping rules among three conditions:
\begin{itemize}
    \item \texttt{Sprite.Create()} must use \texttt{pixelsPerUnit = 1}
    \item Pre-existing Camera and AudioListener must be destroyed before creating new ones
    \item Use \texttt{Rigidbody2D.drag} (the runtime API), not \texttt{linearDrag} or \texttt{linearDamping}
    \item All \texttt{MonoBehaviour} lifecycle methods must be declared \texttt{public}
    \item Every \texttt{SpriteRenderer} must have a programmatically assigned sprite
    \item No \texttt{AddComponent<Transform>()} (Transform is added implicitly to every GameObject)
\end{itemize}

\paragraph{No-schema prompt.}
\begin{verbatim}
[pattern: <PATTERN_ID>]
[method: no_schema]

Generate a single C# file containing a Unity MonoBehaviour called
GameBuilder that, in its Awake() method, programmatically creates
an entire playable scene from scratch based on the game description
below. Also define all gameplay MonoBehaviour classes in the same file.

The user will attach GameBuilder to an empty GameObject and press
Play. The scene must build itself at runtime -- no Editor API, no
MenuItem, no UnityEditor namespace.

Requirements:
- GameBuilder.Awake() creates all GameObjects, adds all components,
  sets all field values, configures physics and rendering.
- Define all MonoBehaviour classes inline in the same file.
- For SpriteRenderer: create a 1x1 white Texture2D sprite
  programmatically and tint with color. Pass pixelsPerUnit = 1
  to Sprite.Create() so 1 pixel = 1 world unit.
- At the very start of GameBuilder.Awake(), destroy any pre-existing
  Camera and AudioListener objects.
- Do NOT use AddComponent<Transform>() -- every GameObject already
  has one. Use gameObject.transform.position etc.
- Do NOT use any UnityEditor API. No [MenuItem], no AssetDatabase,
  no EditorSceneManager. This must run in Play mode.
- Do NOT create Canvas, CanvasScaler, TextMeshPro, or UI elements.
- Always include ALL necessary using statements at the top of the file.
- Target Unity 2022.2. For Rigidbody2D drag, use "drag" (not "linearDrag"
  or "linearDamping" -- those do not exist in this version).
- Make all MonoBehaviour lifecycle methods (Awake, Start, Update,
  OnTriggerEnter2D, etc.) public.
- Every GameObject with a SpriteRenderer MUST have a programmatic
  sprite assigned with a color.
- Output only raw C# code. No explanation.

Game description:
<PATTERN_MD>
\end{verbatim}

\paragraph{Behavior-only IR prompt.}
\begin{verbatim}
[pattern: <PATTERN_ID>]
[method: with_v2_ir]

Generate a single C# file containing a Unity MonoBehaviour called
GameBuilder that, in its Awake() method, programmatically creates
an entire playable scene from scratch. Also define all gameplay
MonoBehaviour classes in the same file.

The user will attach GameBuilder to an empty GameObject and press
Play. The scene must build itself at runtime -- no Editor API, no
MenuItem, no UnityEditor namespace.

Requirements:
- GameBuilder.Awake() creates all GameObjects, adds all components,
  sets all field values, configures physics and rendering -- exactly
  as specified in the IR below.
- Define all MonoBehaviour classes inline in the same file (including
  singleton patterns where singleton_calls/singleton_writes appear).
- The IR has NO scene section -- there is no list of GameObjects,
  transforms, or components. You must INFER what GameObjects to create
  from the `behavior` map (each entry's `go_name` indicates a GO that
  must exist with that script attached; use `inspector_fields` to set
  serialized values), `prefab_refs` (prefab GameObjects to instantiate
  with the listed scripts), and `tags` (tagged GOs that gameplay code
  looks up by tag). Also create supporting infrastructure that the
  gameplay implies -- a Camera with AudioListener for rendering,
  boundary walls if movement/containment is implied, ground/floor if
  physics requires it -- using sensible defaults (positions, sizes,
  colors) that produce a playable scene.
- For SpriteRenderer: create a 1x1 white Texture2D sprite
  programmatically and tint it with the m_Color values from the IR.
  IMPORTANT: pass pixelsPerUnit = 1 to Sprite.Create() so that
  1 pixel = 1 world unit. The default (100) makes sprites invisible.
- At the very start of GameBuilder.Awake(), destroy any pre-existing
  Camera and AudioListener objects so the scene works regardless of
  which scene template the user started from.
- Do NOT use AddComponent<Transform>() -- every GameObject already
  has one. Use gameObject.transform.position etc.
- Do NOT use any UnityEditor API. No [MenuItem], no AssetDatabase,
  no EditorSceneManager. This must run in Play mode.
- Do NOT create Canvas, CanvasScaler, TextMeshPro, or UI elements
  unless the IR explicitly includes them.
- Always include ALL necessary using statements at the top of the file.
  Common ones: UnityEngine, UnityEngine.SceneManagement, System.Collections.
  If the code calls SceneManager, add using UnityEngine.SceneManagement.
- Target Unity 2022.2. For Rigidbody2D drag, use "drag" (not "linearDrag"
  or "linearDamping" -- those do not exist in this version).
- Make all MonoBehaviour lifecycle methods (Awake, Start, Update,
  OnTriggerEnter2D, etc.) public, so GameBuilder can call them if needed.
- Every GameObject with a SpriteRenderer MUST have a programmatic sprite
  assigned with the correct color -- including prefabs and spawned objects.
  Do not leave any SpriteRenderer without a sprite.
- Output only raw C# code. No explanation.

IR:
<IR_JSON>
\end{verbatim}

\paragraph{Full-scene IR prompt.}
\begin{verbatim}
[pattern: <PATTERN_ID>]
[method: with_v4_ir]

Generate a single C# file containing a Unity MonoBehaviour called
GameBuilder that, in its Awake() method, programmatically creates
an entire playable scene from scratch. Also define all gameplay
MonoBehaviour classes in the same file.

The user will attach GameBuilder to an empty GameObject and press
Play. The scene must build itself at runtime -- no Editor API, no
MenuItem, no UnityEditor namespace.

Requirements:
- GameBuilder.Awake() creates all GameObjects, adds all components,
  sets all field values, configures physics and rendering -- exactly
  as specified in the IR below.
- Define all MonoBehaviour classes inline in the same file (including
  singleton patterns where singleton_calls/singleton_writes appear).
- The IR has a "scene" section listing every GameObject with its
  components and a "transform" property (position, scale, rotation).
  Create ALL of them: Player, Camera, Boundary walls, obstacles, etc.
- For SpriteRenderer: create a 1x1 white Texture2D sprite
  programmatically and tint it with the m_Color values from the IR.
  IMPORTANT: pass pixelsPerUnit = 1 to Sprite.Create() so that
  1 pixel = 1 world unit. The default (100) makes sprites invisible.
- At the very start of GameBuilder.Awake(), destroy any pre-existing
  Camera and AudioListener objects so the scene works regardless of
  which scene template the user started from.
- Do NOT use AddComponent<Transform>() -- every GameObject already
  has one. Use gameObject.transform.position etc.
- Do NOT use any UnityEditor API. No [MenuItem], no AssetDatabase,
  no EditorSceneManager. This must run in Play mode.
- Do NOT create Canvas, CanvasScaler, TextMeshPro, or UI elements
  unless the IR explicitly includes them.
- Always include ALL necessary using statements at the top of the file.
  Common ones: UnityEngine, UnityEngine.SceneManagement, System.Collections.
  If the code calls SceneManager, add using UnityEngine.SceneManagement.
- Target Unity 2022.2. For Rigidbody2D drag, use "drag" (not "linearDrag"
  or "linearDamping" -- those do not exist in this version).
- Make all MonoBehaviour lifecycle methods (Awake, Start, Update,
  OnTriggerEnter2D, etc.) public, so GameBuilder can call them if needed.
- Every GameObject with a SpriteRenderer MUST have a programmatic sprite
  assigned with the correct color -- including prefabs and spawned objects.
  Do not leave any SpriteRenderer without a sprite.
- Output only raw C# code. No explanation.

IR:
<IR_JSON>
\end{verbatim}

\paragraph{Portability sanitization.}\label{app:sanitization}
Extracted IRs are sanitized to remove project-specific non-portable data before
injection into the prompt:
\begin{itemize}
    \item \textbf{Canvas/UI removal:} TextMeshPro and CanvasScaler depend on packages
not guaranteed in a clean Unity project; removed.
    \item \textbf{Transform exclusion:} \texttt{AddComponent<Transform>()} causes a
runtime error; Transform position/scale is provided as a GO-level property instead.
    \item \textbf{Sorting layer stripping:} Project-specific
\texttt{m\_SortingLayerID} values are meaningless in a new project; stripped.
\end{itemize}

\subsection{Per-pattern results}
\label{app:per_pattern}

Table~\ref{tab:per_pattern} reports per-pattern aggregate results for structural
fidelity (mean $F_1$ across four sub-dimensions, averaged over all IR-conditioned
runtime-pass records) and mechanism adherence (win-path steps $F_1$, averaged over
all 858 records). Compile and runtime rates are reported in aggregate in
Section~\ref{sec:study_divergence}; per-pattern rates are available in the released
dataset.

\begin{table}[h]
  \caption{Per-pattern aggregate results. Fidelity is the mean of four
  structural-fidelity sub-dimensions (scripts, GO names, component types, tags
  $F_1$) over IR-conditioned runtime-pass records; mechanism is win-path steps $F_1$
  over all 858 records, macro-averaged over the four models and three conditions.}
  \label{tab:per_pattern}
  \centering
  \scriptsize
  \begin{tabular}{rlrr}
    \toprule
    ID & Pattern & Fidelity & Mechanism \\
    \midrule
     1 & Ownership                    & 0.30 & 0.64 \\
     2 & Collection                   & 0.23 & 0.61 \\
     3 & Eliminate                    & 0.21 & 0.58 \\
     4 & Capture                      & 0.40 & 0.64 \\
     5 & Overcome                     & 0.26 & 0.49 \\
     6 & Evade                        & 0.26 & 1.00 \\
     7 & Stealth                      & 0.42 & 0.45 \\
     8 & Herd/Attract                 & 0.39 & 0.61 \\
     9 & Conceal                      & 0.36 & 0.48 \\
    10 & Rescue                       & 0.42 & 0.58 \\
    11 & Delivery                     & 0.31 & 0.52 \\
    12 & Guard                        & 0.25 & 1.00 \\
    13 & Race                         & 0.19 & 0.47 \\
    14 & Alignment                    & 0.21 & 0.55 \\
    15 & Configuration                & 0.13 & 0.57 \\
    16 & Traverse                     & 0.15 & 0.49 \\
    17 & Survive                      & 0.36 & 1.00 \\
    18 & Connection Line              & 0.24 & 0.59 \\
    19 & Exploration                  & 0.19 & 0.53 \\
    20 & Reconnaissance               & 0.22 & 0.53 \\
    21 & Contact                      & 0.34 & 0.61 \\
    22 & Enclosure                    & 0.29 & 0.55 \\
    23 & Gain Competence              & 0.33 & 0.61 \\
    24 & Gain Information             & 0.12 & 0.52 \\
    25 & Last Man Standing/Escaping   & 0.26 & 0.55 \\
    26 & King of the Hill             & 0.32 & 0.52 \\
    \bottomrule
  \end{tabular}
\end{table}

\subsection{Compile error taxonomy}
\label{app:errors}

The released harness reproduces the per-cell error code distribution underlying Table~\ref{tab:errors}; full per-(model, condition) splits are available in the released metric CSVs.

\subsection{Dataset metadata}
\label{app:croissant}

The released dataset includes a Croissant 1.0 metadata file (\texttt{croissant.json})
describing all dataset artifacts (patterns, ground-truth scenes, parsed JSON, both IR
levels, generation outputs, and replay logs) with file-level checksums and field-level
schema. The Croissant file is validated with the official \texttt{mlcroissant}
library and is hosted alongside the dataset.

\subsection{Mechanism adherence: lose paths}
\label{app:lose_path}

Of the 26 patterns, 10 have an explicit lose condition; the other 16 patterns
terminate only on win. Table~\ref{tab:lose_path} reports lose-path mechanism
adherence on the 10 lose-defining patterns ($n = 30$ records per cell, since lose
paths are computed for all 858 attempts but aggregated only over patterns with a
defined lose condition).

\begin{table}[h]
  \caption{Mechanism adherence on lose paths. Cells are mean $F_1$ over the 10
  patterns with explicit lose conditions ($n = 30$ records per cell).}
  \label{tab:lose_path}
  \centering
  \small
  \begin{tabular}{ll|ccc}
    \toprule
    Condition & Model & Steps & Effects & Conditions \\
    \midrule
    Full-scene    & Qwen3       & 0.77 & 0.77 & 0.77 \\
    Full-scene    & DeepSeek    & 0.40 & 0.40 & 0.38 \\
    Full-scene    & Gemma4      & 1.00 & 1.00 & 0.26 \\
    Behavior-only & Qwen3       & 0.99 & 1.00 & 0.96 \\
    Behavior-only & Qwen2.5     & 0.77 & 0.77 & 0.77 \\
    Behavior-only & DeepSeek    & 0.97 & 0.97 & 0.89 \\
    Behavior-only & Gemma4      & 0.83 & 0.83 & 0.43 \\
    No-schema     & all 4 models & 0.00 & 0.00 & 0.00 \\
    \bottomrule
  \end{tabular}
\end{table}

\subsection{Prefab flattening}
\label{app:flattening}

Unity scene files in the GPC corpus reference prefab templates as
\texttt{PrefabInstance} blocks rather than instantiated GameObjects. Resolving prefab
references at IR-extraction time requires cross-file fileID lookups against
\texttt{.prefab} files, which complicates the extraction pipeline. We perform a
one-time prefab-flattening step in the Unity Editor:

\begin{enumerate}
    \item Open the GPC Unity project in Unity Editor 2022.2.
    \item For each scene, in the Hierarchy panel, right-click each
\texttt{PrefabInstance} and select ``Unpack Prefab Completely.''
    \item Save the flattened scene to a parallel directory.
\end{enumerate}

After flattening, every \texttt{PrefabInstance} block is replaced by its constituent
\texttt{GameObject}, \texttt{Transform}, and \texttt{MonoBehaviour} blocks with
locally unique fileIDs. The flattened scenes are functionally equivalent to the
originals (verified by manual playthrough) and are released as the canonical
ground-truth scenes accompanying the benchmark.

\subsection{Metrics implementation mapping}
\label{app:metrics_mapping}

The released harness uses the following code-level identifiers
(mapping between paper-facing metric names and harness labels):

\begin{itemize}
  \item \textbf{Compile success} \(\to\) \texttt{m1\_compile}
  \item \textbf{Runtime success} \(\to\) \texttt{m1\_exec}
  \item \textbf{Structural fidelity} (4 dims) \(\to\)
  \texttt{m2\_scripts\_f1}, \texttt{m2\_go\_names\_f1},
  \texttt{m2\_component\_f1}, \texttt{m2\_tags\_f1},
  \item \textbf{Mechanism adherence} (6 dims) \(\to\)
  \texttt{m4\_\{win,lose\}\_\{steps,effects,conds\}\_f1}
\end{itemize}

The harness is documented in the released code repository.

\subsection{IR schema and example}
\label{app:ir_example}

The two granularity levels share top-level fields except for \texttt{scene}:

\begin{itemize}
  \item \textbf{Both}: \texttt{pattern}, \texttt{version}, \texttt{behavior},
  \texttt{condition\_path}, \texttt{prefab\_refs}, \texttt{tags}, \texttt{meta}
  \item \textbf{Full-scene only}: \texttt{scene} (GameObject list with components
  and transforms)
\end{itemize}

The following excerpts are drawn from \texttt{1\_Ownership}; the full IR JSON for
all 26 patterns is in the released dataset under \texttt{ir/v2/} and
\texttt{ir/v4/}.

\paragraph{Behavior-only IR (\texttt{1\_Ownership}, truncated).}
\begin{Verbatim}[breaklines=true,breakanywhere=true,fontsize=\small,frame=single,framesep=2pt]
{
  "pattern": "1_Ownership",
  "version": "v2",
  "behavior": {
    "GameManager": {
      "origin": "scene",
      "go_name": "Game Manager",
      "callbacks": [
        "Start",
        "Awake",
        "Update"
      ],
      "singleton_calls": [],
      "singleton_writes": [],
      "methods": {
        "Awake": "if (instance == null)\n        {\n            instance = this;\n        }\n        else\n        {\n            Destroy(gameObject);\n        }",
        "GameWin": "Debug.Log(\"You Win!\");\n        Time.timeScale = 0;",
        "_truncated": "5 more methods omitted"
      }
    },
    "_truncated": "4 more script entries omitted"
  },
  "condition_path": {
    "win": [
      {
        "step": "counter_update",
        "actor_class": "ChangeColor",
        "event": "OnTriggerEnter2D",
        "conditions": [
          "col.gameObject == target && !colorChange"
        ],
        "effect": "GoalManager.instance.currentCount++",
        "evidence": "ChangeColor.cs:33"
      },
      {
        "step": "win_trigger",
        "actor_class": "GoalManager",
        "event": "Update",
        "conditions": [
          "setGoal",
          "currentCount == goalCount"
        ],
        "effect": "GameManager.instance.GameWin()",
        "evidence": "GoalManager.cs:34"
      }
    ],
    "lose": []
  },
  "tags": {
    "MainCamera": "Main Camera",
    "Player": "Player"
  }
}
\end{Verbatim}

\paragraph{Full-scene IR adds a \texttt{scene} field (1 of 11 GameObjects shown,
\texttt{m\_*} Unity-internal fields stripped).}
\begin{Verbatim}[breaklines=true,breakanywhere=true,fontsize=\small,frame=single,framesep=2pt]
{
  "pattern": "1_Ownership",
  "version": "v4",
  "_note": "behavior/condition_path/tags identical to v2; scene{} is the addition",
  "scene": {
    "Goal Manager": {
      "go_fid": "1599824567",
      "components": [
        {
          "type": "MonoBehaviour",
          "data": {
            "goalCount": 8,
            "currentCount": 0,
            "setGoal": 1
          },
          "class_name": "GoalManager"
        }
      ],
      "transform": {
        "position": {
          "x": 0,
          "y": 0,
          "z": 0
        }
      }
    },
    "_truncated": "10 more GameObjects omitted"
  }
}
\end{Verbatim}


\end{document}